\documentclass{article}
\usepackage{spconf,amsmath,graphicx}

\usepackage{cite}
\usepackage{amsmath,amssymb,amsfonts}
\usepackage{algorithmic}
\usepackage{graphicx}
\usepackage{textcomp}
\usepackage{xcolor}
\usepackage{hyperref}
\usepackage{here}
\usepackage{array}
\usepackage{subcaption}
\usepackage{cleveref} 

\DeclareMathOperator{\SL}{SL}

\def\BibTeX{{\rm B\kern-.05em{\sc i\kern-.025em b}\kern-.08em
    T\kern-.1667em\lower.7ex\hbox{E}\kern-.125emX}}
\begin{document}

\title{StampNet: unsupervised multi-class object discovery}

\name{Joost Visser \qquad Alessandro Corbetta \qquad Vlado Menkovski \qquad Federico Toschi}
\address{Eindhoven University of Technology}

\maketitle

\begin{abstract}
Unsupervised object discovery in images involves uncovering recurring patterns that define objects and discriminates them against the background. This is more challenging than image clustering as the size and the location of the objects are not known: this adds additional degrees of freedom and increases the problem complexity. In this work, we propose StampNet, a novel autoencoding neural network that localizes shapes (objects) over a simple background in images and categorizes them simultaneously. StampNet consists of a discrete latent space that is used to categorize objects and to determine the location of the objects. The object categories are formed during the training, resulting in the discovery of a fixed set of objects. We present a set of experiments that demonstrate that StampNet is able to localize and cluster multiple overlapping shapes with varying complexity including the digits from the MNIST dataset. We also present an application of StampNet in the localization of pedestrians in overhead depth-maps.
\end{abstract}


\section{Introduction}
Discovery and localization of objects is an important task in computer vision and image analysis. There is a significant amount of existing methods that successfully address the challenge of object localization when objects are explicitly annotated with labels and bounding boxes.  Deep convolutional neural networks such as YOLO~\cite{redmon2018yolov3} and Faster R-CNN~\cite{ren2015faster} demonstrated significant success. Annotations, however, often require major human effort and thus come with significant costs. 

In contrast, object discovery deals with localization in absence of annotations. This means finding and clustering recurring patterns that define the objects. Previous work on unsupervised localization typically addresses one object class in multiple images, which is referred to as object co-localization~\cite{kim2009unsupervised,tang2014co}. However, unsupervised localization of multiple classes of objects remains a significant challenge. This problem is further adds complexity as we cannot assume that only a single object is present in the image. In other words, the model needs to simultaneously discover the objects (or form categories) and learn to perform localization.

As the object size and alignment is not predetermined the complexity of the clustering of the objects grows significantly with the degrees of freedom added by the localization. Analogously, the localization is difficult because during training objects categories are not predetermined. An additional complication is that the objects can overlap, which further increases the difficulty of clustering the objects. 

To address these challenges we propose a novel autoencoding neural network architecture, StampNet, that discovers objects and localizes them simultaneously. StampNet has two characterizing features: first, it has a latent space consisting of discrete random variables that encode the cluster assignment and its location. Second, it has a final layer consisting of a fixed number of convolutional filters (\textit{stamps}) that encode the discovered objects. The size and number of these filters determine the maximum size and maximum number of objects to be discovered respectively. We refer to this layer as a \textit{stamp layer} and hence the name of the network StampNet. 

This paper is structured as follows: in~\Cref{sec:related-works}, we report on the state-of-the-art on unsupervised object discovery. In~\Cref{sec:architecture}, we introduce the StampNet architecture. In~\Cref{sec:loc-results}, we show the results of StampNet on four datasets. The discussion~\Cref{sec:conclusion} closes the paper.

\section{Related works} \label{sec:related-works}

Various studies have been performed on topics closely related to unsupervised object discovery. In co-localization and co-segmentation, the goal is to extract the position of common objects between images, using bounding boxes and segmentation respectively. Many of these studies obtain good results, but simplify the problem by assuming a single object class ~\cite{kim2009unsupervised,tang2014co,rubinstein2013unsupervised}. 

Recent studies have worked on the more difficult problem of multi-class co-localization~\cite{wang2014unsupervised} and co-segmentation~\cite{cho2015unsupervised}. Cho et al.~\cite{cho2015unsupervised} use an off-the-shelf region proposal system to form a set of candidate bounding boxes and match these across images. Wang et al.~\cite{wang2014unsupervised} use functional maps to model partial similarity across images, but they assume that the image set contains two classes of objects or very similar objects. 

To the best of our knowledge, only one study has focussed on our task of multi-class object discovery: tackling simultaneous localization and classification without supervision. Murasaki et al.~\cite{murasaki2019unsupervised} extract deep features using a pre-trained neural network and clusters these features. However, their model is limited to a single single object in each image.

\section{StampNet architecture} \label{sec:architecture}
\subsection{Architecture overview}
\begin{figure}[htbp]
  \centerline{\includegraphics[width=\linewidth]{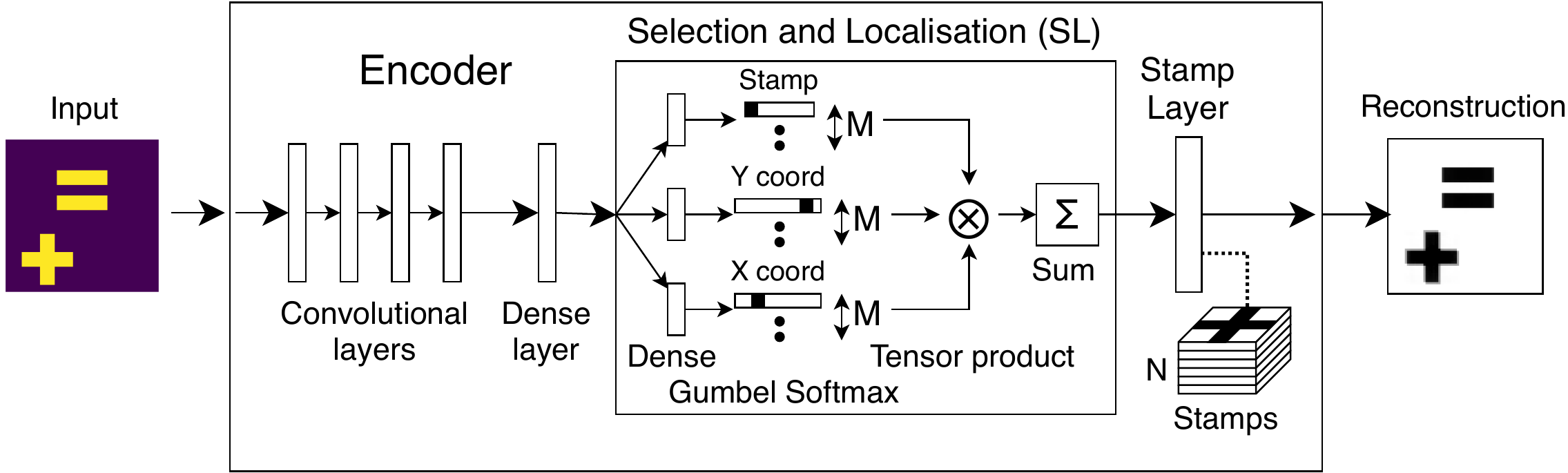}}
  \caption{The StampNet architecture consists of an encoder (convolutional neural network), selection-and-localization layer and the stamp layer. The encoder produces a map from image space to the latent space. The selection-and-localization layer maps the latent representation to an activation map for the stamp layer (\Cref{sec:SL-layer}). Finally, the stamp layer produces the reconstruction of the image based on the activation map (\Cref{sec:stamp-layer}).
  \vspace*{-3.9mm}
  }
  \label{fig:architecture}
\end{figure}

In this section, we describe the architecture of StampNet (see sketch  in~\autoref{fig:architecture}). StampNet is an autoencoder, i.e. a neural network which outputs a reconstruction of an input image based on an internal representation of the input. This representation is formed by the encoder under the training constraint of minimising Euclidean distance between the input and the output. 

The encoder follows a VGG-like~\cite{simonyan2014very} structure with stacks of convolutional layers (with Leaky ReLU activations~\cite{maas2013rectifier}) followed by a max pooling layer (Batch Normalization~\cite{ioffe2015batch} and Dropout~\cite{srivastava2014dropout} are used during training). %

The decoder, the peculiarity of StampNet, works in two stages: 
first, it predicts the coordinates and stamp type for each  shape in the input image (Selection-and-localization layer); 
second, through the stamp layer, the selected stamps are placed according to the predicted positions to generate the output image. In the following subsection, we introduce the notation employed and we detail the selection-and-localization and the stamp layers. 

\subsection{Notation}
We consider an input image of pixel size $\phi_x \times\phi_y$ containing $m \leq M$ shapes indexed by the variable $\eta$ (i.e. $1\leq \eta \leq M$, where the upper-bound $M$ to the number of shapes is pre-defined). We assume each shape to fit within one of $N$ stamps of pixel size $\psi_x \times \psi_y$. The set of stamps is stored in a collection of $N$ convolutional kernels of size $\psi_x \times \psi_y$, that form, as whole, a tensor  $\Omega$ of size $N \times \psi_x \times \psi_y$.
We finally name $X$ and $Y$ the random variables respectively associated to the $x$ and $y$ coordinate of a shape and $S$ the categorical random variable defining the stamp type $s$. Thus, it holds $0\leq x,y \leq \phi_x - \psi_x + 1,\phi_y - \psi_y + 1 $ and $0\leq s \leq N-1$.

\subsection{Selection-and-localization layer}\label{sec:SL-layer}
The selection-and-localization layer (SL-layer) predicts the coordinates (localization) and selects the clustered object to use (selection). As objects are clustered in the kernel of the stamp layer, these need to be convolved over the predicted location and only the selected object should be convolved. This happens when the output of the SL-layer is a one-hot tensor, (see \autoref{eq:stamp}).

Therefore, we want to output a one-hot tensor at the predicted coordinates and predicted stamp. One way to output such tensor is to use a max function. This function, however, results in a gradient of $0$ for all non-max values, making it unfitting for backpropagation. We provide a different solution: we model the coordinates and object-selector as categorical distributions. When we sample from these distributions, we get a one-hot tensor with the predicted stamp and coordinates. The \textit{gumbel softmax}~\cite{jang2016categorical} (also known as the \textit{concrete distribution}~\cite{maddison2016concrete}) is a differentiable sampling of a smoothly deformed categorical distribution.

In the SL-layer, we first predict the probability distribution of the $x$ coordinate, $y$ coordinate and the stamp type $s$. Then, we calculate a sampling using the gumbel softmax function. This function uses temperature $\tau$ that allows gradients to flow through during training and gradually anneals the temperature to become a sampling from a categorical distribution~\cite{jang2016categorical}.

The network combines the three gumbel softmax output by performing, for each of the $M$ shapes, the tensor product of the coordinates and of the stamp selector probability vectors. We obtain $M$ individual selection-and-localization tensors (SL tensors), as they contain information on the position and type of each shape. In formulae, the individual SL tensor reads for all $1 \leq \eta \leq M$:  
\begin{align}\label{eq:sl-layer}
    SL_{ijk}^{(\eta)} = P(X^{(\eta)}=i)P(Y^{(\eta)}=j)P(S^{(\eta)}=k) %
\end{align}

Once every shape has been selected and localized, we combine them into the global selection-and-localization tensor summing on $\eta$:
\[
\SL_{ijk} = \sum_{\eta=1}^M SL_{ijk}^{(\eta)}.
\]

\subsection{The stamp layer}\label{sec:stamp-layer}
The output $O$ of the network, i.e. the reconstruction of the input, is provided by the stamp layer, which performs a convolution operation between the global $\SL$ tensor and the stamp tensor $\Omega$. We write $O = \SL \ast \Omega$, which in components satisfies: %
\begin{align} \label{eq:stamp}
    O_{i^*j^*} 
     &= \sum_{k=1}^N \left( \sum_{ij} \SL_{ijk} \Omega_{(i^* - i) (j^* -j) k}\right)
\end{align}

We constrain the kernel values in $\Omega$ to be non-negative and smaller than the maximum image value with training-time clipping. Moreover, should two stamps be partially overlapped, and thus sum their values, the output is further clipped to avoid exceeding the maximum image value.

\section{Results} \label{sec:loc-results}

\begin{figure*}[ht]
  \centering
  \begin{subfigure}[b]{0.33\linewidth}
    \centering\includegraphics[width=\linewidth]{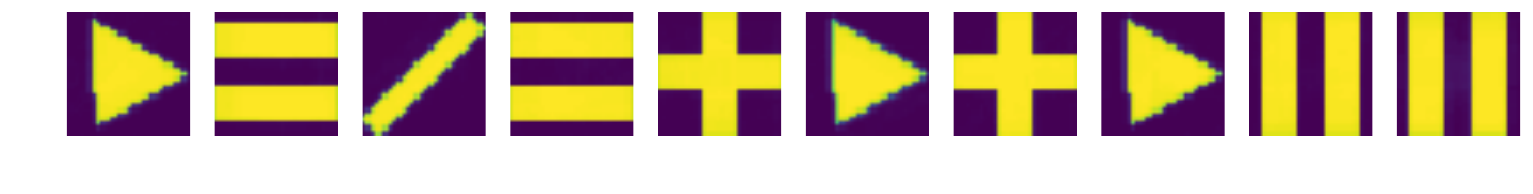}
    \vspace*{-7.2mm}
    \caption{}
    \label{fig:simple-shapes-stamps}
  \end{subfigure}
  \hfill
  \begin{subfigure}[b]{0.66\linewidth}
    \centering\includegraphics[width=\linewidth]{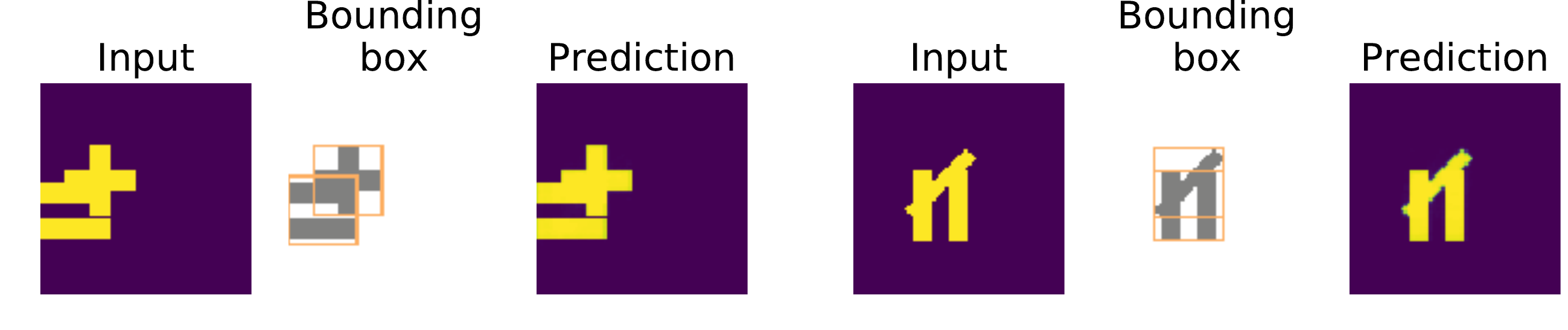}
    \vspace*{-7.2mm}
    \caption{}
    \label{fig:simple-shapes-results}
  \end{subfigure}

  \begin{subfigure}[b]{0.33\linewidth}
    \centering\includegraphics[width=\linewidth]{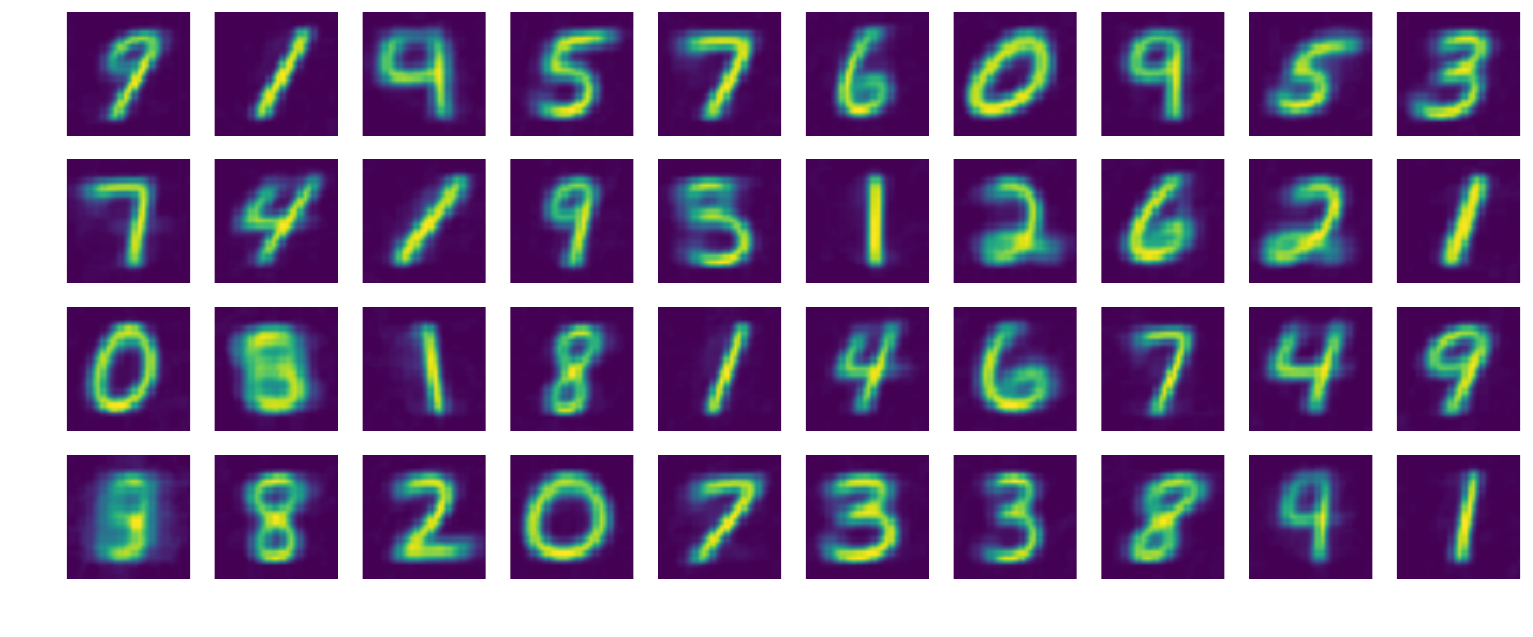}
    \vspace*{-7.2mm}
    \caption{}
    \label{fig:t-mnist-stamps}
  \end{subfigure}
  \hfill
  \begin{subfigure}[b]{0.66\linewidth}
    \centering\includegraphics[width=\linewidth]{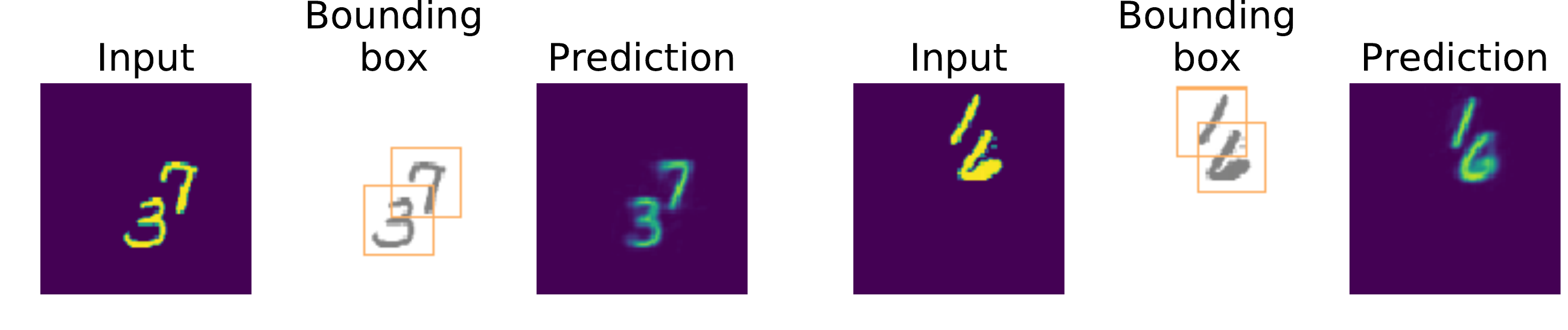}
    \vspace*{-7.2mm}
    \caption{}
    \label{fig:t-mnist-results}
  \end{subfigure}

  \begin{subfigure}[b]{0.33\linewidth}
    \centering\includegraphics[width=\linewidth]{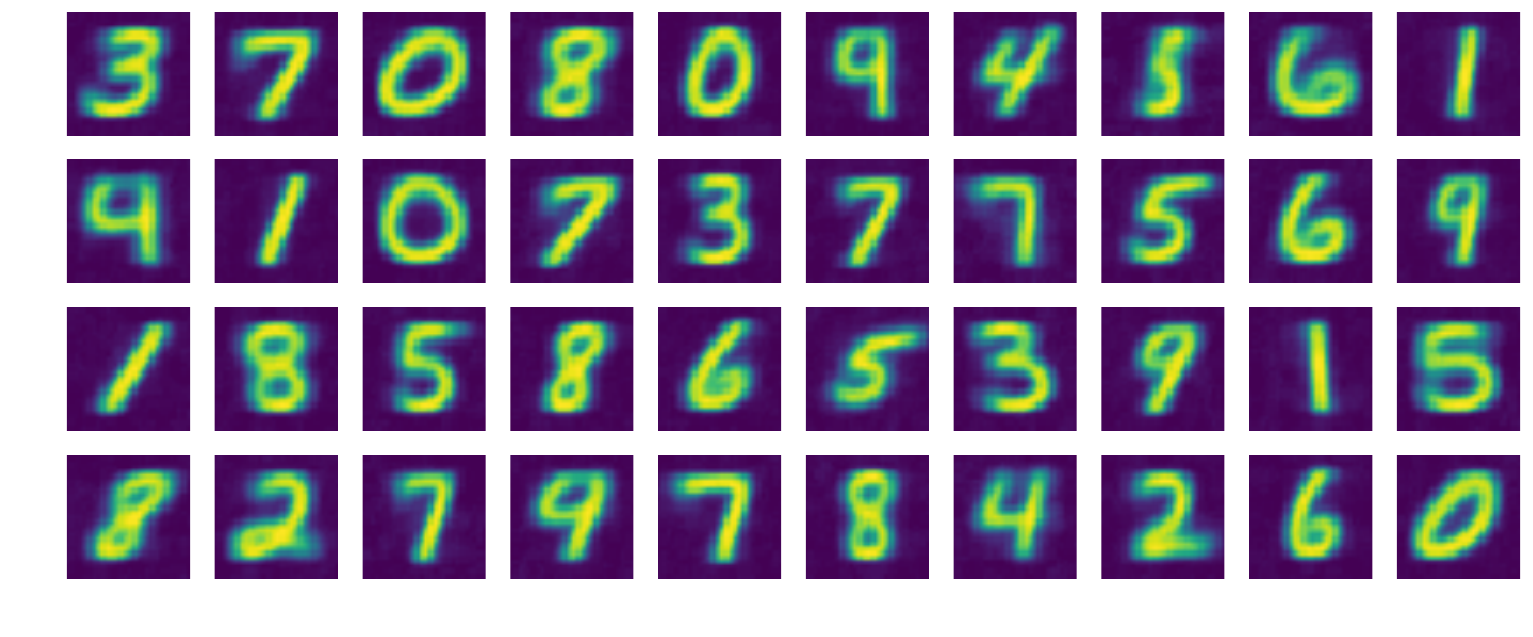}
    \vspace*{-7.2mm}
    \caption{}
    \label{fig:ct-mnist-stamps}
  \end{subfigure}
  \hfill
  \begin{subfigure}[b]{0.66\linewidth}
    \centering\includegraphics[width=\linewidth]{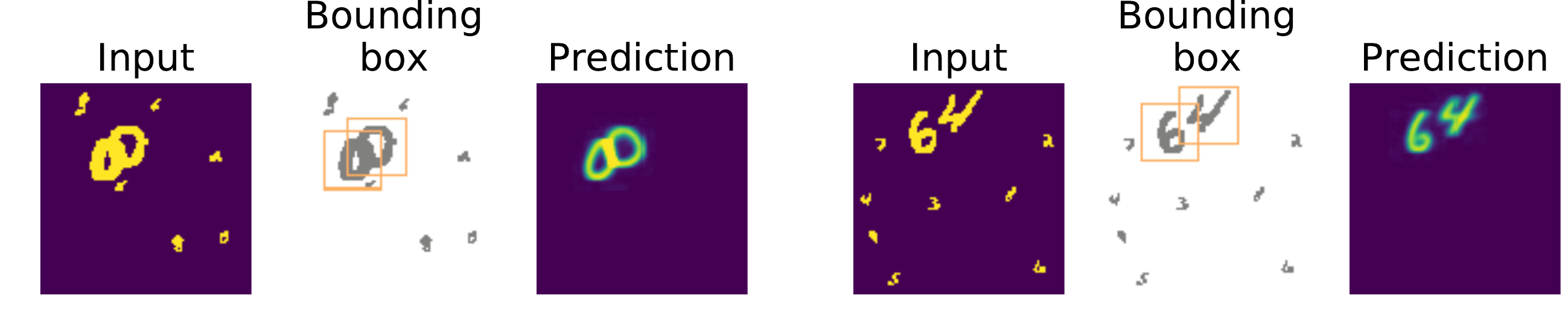}
    \vspace*{-7.2mm}
    \caption{}
    \label{fig:ct-mnist-results}
  \end{subfigure}
  \begin{subfigure}[b]{0.33\linewidth}
    \centering\includegraphics[width=\linewidth]{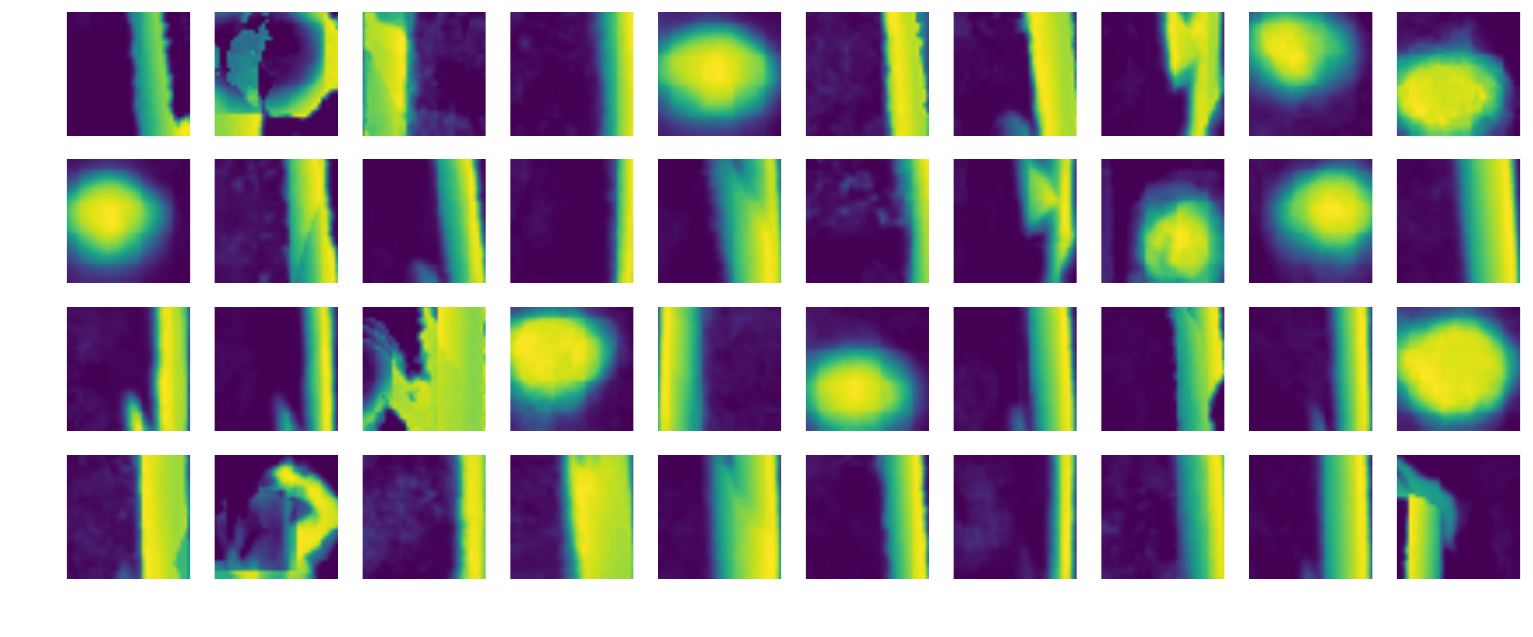}
    \vspace*{-7.2mm}
    \caption{}
    \label{fig:ped-stamps}
  \end{subfigure}
  \hfill
  \begin{subfigure}[b]{0.66\linewidth}
    \centering\includegraphics[width=\linewidth]{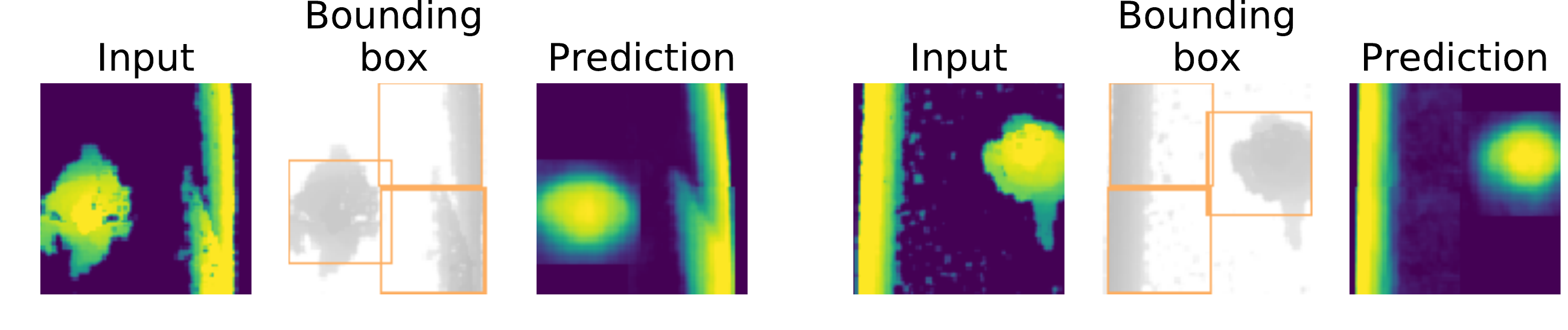}
    \vspace*{-7.2mm}
    \caption{}
    \label{fig:ped-results}
  \end{subfigure}
  \caption{ The results of StampNet on four different datasets: Simple Shapes dataset (a,b), Translated MNIST (c,d), Cluttered Translated MNIST (e,f) and Pedestrian Tracking dataset (g,h). 
  \textit{Left:} (a,c,e,g) stamps learned by our model in the Stamp Layer (see \autoref{fig:architecture}) \textit{Right:} (b,d,f,h) samples of each dataset . The network predicts bounding boxes (orange) and places a stamp at these locations to reconstruct the input image as close as possible. 
  }
\end{figure*}

We evaluate the performance of StampNet on several datasets. The first dataset contains five clearly distinguishable shapes, followed by more complex shapes of MNIST. The final dataset contains noisy overhead images of pedestrians.  The performance of the network is evaluated on two tasks: (1) discovery and localization of each shape and (2) clustering performance of the extracted stamps. We evaluate the first task with CorLoc~\cite{cho2015unsupervised} and Intersection over Union (IoU)~\cite{everingham2010pascal}. For the second task, we use clustering purity~\cite{yang2012clustering} to measure how well the model can differentiate between classes. %

The network has been trained on a train set and evaluated on a separate test set. We use an annealing schedule of $\tau = \max (0.2, 7.0\cdot \exp{(-0.01 \cdot t)})$, updated every epoch for the gumbel softmax. For the results, we use a temperature of $\tau=0.01$ to enforce a one-hot choice.

\subsection{Simple Shapes dataset}

The Simple Shapes dataset uses five different simple shapes: a `plus', an `equal', a rotated `equal', a `slash',  and a `triangle'.
We randomly place $M=2$ shapes of size $28 \times 28$ on an empty canvas of size $84 \times 84$. We consider $50000$ generated training samples and $10000$ generated test samples. The stamp layer contains $N=10$ stamps.

In \autoref{fig:simple-shapes-stamps}, we report the $N=10$ stamps learned by the network. We observe that these stamps are nearly identical to the shapes used to generate the dataset. We detail sample predictions outputs in \autoref{fig:simple-shapes-results}. In these samples, the network localizes overlapping shapes and assigns the correct stamp to each shape.

\begin{table}[htbp]
    \caption{Experiment results \vspace{-2mm}}
    \begin{center}
        \begin{tabular}{c|c|c|c}
            \textbf{Dataset} & \textbf{CorLoc} & \textbf{IoU} & \textbf{Purity} \\ \hline
            Simple Shapes $(M=1)$ & 0.9999 & 0.9718 & 0.9928 \\  
            Simple Shapes $(M=2)$ & 0.9828 & 0.9500 & 0.9564 \\  
            T-MNIST $(M=1)$ & 0.9983 & 0.8925 & 0.7891 \\
            T-MNIST $(M=2)$  & 0.9729 & 0.8537 & 0.5277 \\  
            CT-MNIST $(M=1)$ & 0.9972 & 0.8912 & 0.8113 \\
            CT-MNIST $(M=2)$ & 0.9545 & 0.8394 & 0.6149 \\ 	
            Pedestrian Tracking $(M=3)$ & 0.7816$^{\mathrm{a}}$ & 0.6308$^{\mathrm{a}}$ & 0.9597 \\[0.44mm]
            \multicolumn{4}{l}{$^{\mathrm{a}}$Calculated for localizing the pedestrian, not the walls.}
            \vspace{-3mm}
        \end{tabular}
        \label{tab:metrics}
    \end{center}
\end{table}

\begin{table}[htbp]
    \caption{Supervised classification CT-MNIST $(M=1)$ \vspace{-2mm}}
    \begin{center}
    \begin{tabular}{c|c}
        \textbf{Network} & \textbf{Accuracy} \\ \hline
        RAM~\cite{mnih2014recurrent} & 0.927 \\
        DRAW~\cite{gregor2015draw} & 0.966 \\  
        RNN-SPN~\cite{sonderby2015recurrent}& 0.985 \\
        DCN~\cite{almahairi2016dynamic} & 0.986 \vspace{-5mm}

    \end{tabular}
    \label{tab:clutter-supervised}
    \end{center}
\end{table}

In \autoref{tab:metrics}, we quantify the IoU and purity of evaluated over the test set. We achieve high scores in IoU and purity on this dataset. The remaining errors are the result of the network assigning an inaccurate stamp to a shape, which primarily happens when shapes overlap.

\subsection{Translated and Cluttered Translated MNIST}
We evaluate the performance of our model on two MNIST datasets: (1) Translated MNIST (T-MNIST) and (2) Cluttered Translated MNIST (CT-MNIST) \cite{mnih2014recurrent}. In T-MNIST, MNIST digits are uniformly placed on an empty canvas. CT-MNIST adds clutter to these images by uniformly placing $8$ smaller clutter digits of size $8 \times 8$ to add noise to the dataset. We choose in T-MNIST a canvas size of $84 \times 84$ and in CT-MNIST a canvas size of $100 \times 100$ to best compare our results with existing literature. In both cases, we generate $60 000$ training samples and $10 000$ test samples. We test for $M=1$ and $M=2$ digits on the canvas with $N=40$ stamps.

We observe that the network discovers different MNIST digits in \autoref{fig:t-mnist-results} and \ref{fig:ct-mnist-results}. The stamps learned by the network (\autoref{fig:t-mnist-stamps} and \ref{fig:ct-mnist-stamps}) resemble different digits of MNIST. 

The network discovers most MNIST digits, as the CorLoc measures (\autoref{tab:metrics}) indicate. Even when there are $M=2$ digits on a cluttered canvas, the network discovers over 95\% of the digits. The added clutter results in a slight drop of localization (\autoref{tab:metrics}) and an increase in purity.

We note the results of others in the case of CT-MNIST $(M=1)$ in \autoref{tab:clutter-supervised}. We observe that without supervision, StampNet performance comes near these supervised alternatives (using purity as the measure for comparison, i.e when all stamps are labelled correctly). Note that these networks only classify a single digit, while our model can discover multiple digits in each image. 

\subsection{Pedestrian localization in overhead images} \label{sec:ped-track}

Overhead depth maps are an increasingly popular tool to perform high accuracy pedestrian tracking for studying the dynamics of human crowds in real-life venues  (e.g.~\cite{seer2014kinects}). %
Overhead depth maps come in form of grey scale images where the value of each pixel encodes its distance from the camera plane. In overhead depth view pedestrians have similar ``ovoid'' shape, which is different from that of walls, objects and so on. The characteristics of this dataset make it a very suitable for the StampNet's object discovery capability. 

We consider here a reduced depth dataset from the real-life crowd tracking experiment~\cite{corbetta2018large}, annotated with bounding boxes (image size: $80 \times 80$, bounding boxes size $40\times 40$. See sample  on the left side of~\autoref{fig:ped-results}, which displays a pedestrian on the left side and a wall on the right side).

We test StampNet considering $N=40$ stamps of size $40 \times 40$ and evaluate both the accuracy of the localization and the capability of the network to differentiate between pedestrians and the wall on the side.

We illustrate samples of the results in~\autoref{fig:ped-results} and the measures in~\autoref{tab:metrics}. In both cases, we observe that the network places one stamp on the pedestrian and two stamps on the walls. The stamps extracted by the network (\autoref{fig:ped-stamps}) show that different objects are successfully captured by the network, and, as evidenced by the high clustering purity, the network is capable to differentiate between pedestrians and walls. Furthermore, we obtain a CurLoc value of $0.78$ showing that we can localize with reasonable accuracy.

\section{Conclusion} \label{sec:conclusion}
In this paper, we have introduced StampNet to localize multiple objects from multiple classes in an unsupervised manner. We accomplish this by incorporating the predictions of the bounding boxes into an autoencoder, removing the need of labels. We achieve this by placing the kernel of a convolutional layer on predicted location coordinates (\autoref{fig:architecture}).

The results in~\autoref{fig:t-mnist-results} and \ref{fig:ct-mnist-results} show that StampNet is able to detect and localize overlapping MNIST digits without the need for any labels. Furthermore, the network clusters the shapes in the dataset as stamps (\autoref{fig:ct-mnist-stamps}). We demonstrate an example of the value of this in an application of pedestrian tracking in overhead images. 

Nevertheless, there are limitations to our model. The network is only able to place stamps of a fixed size. Furthermore, as we make use of the kernel of a convolutional layer, the network can only capture simple prototypical shapes. 

Future work can look into generating more complex kernels for the stamp layer by making use of information from the input.

\bibliographystyle{IEEEbib}
\bibliography{refs}

\begin{thebibliography}{10}

\bibitem{redmon2018yolov3}
Joseph Redmon and Ali Farhadi,
\newblock ``Yolov3: An incremental improvement,''
\newblock {\em arXiv preprint arXiv:1804.02767}, 2018.

\bibitem{ren2015faster}
Shaoqing Ren, Kaiming He, Ross Girshick, and Jian Sun,
\newblock ``Faster r-cnn: Towards real-time object detection with region
  proposal networks,''
\newblock in {\em Advances in neural information processing systems}, 2015, pp.
  91--99.

\bibitem{kim2009unsupervised}
Gunhee Kim and Antonio Torralba,
\newblock ``Unsupervised detection of regions of interest using iterative link
  analysis,''
\newblock in {\em Advances in neural information processing systems}, 2009, pp.
  961--969.

\bibitem{tang2014co}
Kevin Tang, Armand Joulin, Li-Jia Li, and Li~Fei-Fei,
\newblock ``Co-localization in real-world images,''
\newblock in {\em Proceedings of the IEEE conference on computer vision and
  pattern recognition}, 2014, pp. 1464--1471.

\bibitem{rubinstein2013unsupervised}
Michael Rubinstein, Armand Joulin, Johannes Kopf, and Ce~Liu,
\newblock ``Unsupervised joint object discovery and segmentation in internet
  images,''
\newblock in {\em Proceedings of the IEEE conference on computer vision and
  pattern recognition}, 2013, pp. 1939--1946.

\bibitem{wang2014unsupervised}
Fan Wang, Qixing Huang, Maks Ovsjanikov, and Leonidas~J Guibas,
\newblock ``Unsupervised multi-class joint image segmentation,''
\newblock in {\em Proceedings of the IEEE Conference on Computer Vision and
  Pattern Recognition}, 2014, pp. 3142--3149.

\bibitem{cho2015unsupervised}
Minsu Cho, Suha Kwak, Cordelia Schmid, and Jean Ponce,
\newblock ``Unsupervised object discovery and localization in the wild:
  Part-based matching with bottom-up region proposals,''
\newblock in {\em Proceedings of the IEEE Conference on Computer Vision and
  Pattern Recognition}, 2015, pp. 1201--1210.

\bibitem{murasaki2019unsupervised}
Kazuhiko Murasaki, Yukinobu Taniguchi, and Tetsuya Kinebuchi,
\newblock ``Unsupervised multi-class object discovery by spherical clustering
  of deep features,''
\newblock {\em ITE Transactions on Media Technology and Applications}, vol. 7,
  no. 1, pp. 2--10, 2019.

\bibitem{simonyan2014very}
Karen Simonyan and Andrew Zisserman,
\newblock ``Very deep convolutional networks for large-scale image
  recognition,''
\newblock {\em arXiv preprint arXiv:1409.1556}, 2014.

\bibitem{maas2013rectifier}
Andrew~L Maas, Awni~Y Hannun, and Andrew~Y Ng,
\newblock ``Rectifier nonlinearities improve neural network acoustic models,''
\newblock in {\em International Conference on Machine Learning}, 2013, vol.~30,
  p.~3.

\bibitem{ioffe2015batch}
Sergey Ioffe and Christian Szegedy,
\newblock ``Batch normalization: Accelerating deep network training by reducing
  internal covariate shift,''
\newblock {\em arXiv preprint arXiv:1502.03167}, 2015.

\bibitem{srivastava2014dropout}
Nitish Srivastava, Geoffrey Hinton, Alex Krizhevsky, Ilya Sutskever, and Ruslan
  Salakhutdinov,
\newblock ``Dropout: a simple way to prevent neural networks from
  overfitting,''
\newblock {\em The Journal of Machine Learning Research}, vol. 15, no. 1, pp.
  1929--1958, 2014.

\bibitem{jang2016categorical}
Eric Jang, Shixiang Gu, and Ben Poole,
\newblock ``Categorical reparameterization with gumbel-softmax,''
\newblock {\em arXiv preprint arXiv:1611.01144}, 2016.

\bibitem{maddison2016concrete}
Chris~J Maddison, Andriy Mnih, and Yee~Whye Teh,
\newblock ``The concrete distribution: A continuous relaxation of discrete
  random variables,''
\newblock {\em arXiv preprint arXiv:1611.00712}, 2016.

\bibitem{everingham2010pascal}
Mark Everingham, Luc Van~Gool, Christopher~KI Williams, John Winn, and Andrew
  Zisserman,
\newblock ``The pascal visual object classes (voc) challenge,''
\newblock {\em International journal of computer vision}, vol. 88, no. 2, pp.
  303--338, 2010.

\bibitem{yang2012clustering}
Zhirong Yang, Tele Hao, Onur Dikmen, Xi~Chen, and Erkki Oja,
\newblock ``Clustering by nonnegative matrix factorization using graph random
  walk,''
\newblock in {\em Advances in Neural Information Processing Systems}, 2012, pp.
  1079--1087.

\bibitem{mnih2014recurrent}
Volodymyr Mnih, Nicolas Heess, Alex Graves, et~al.,
\newblock ``Recurrent models of visual attention,''
\newblock in {\em Advances in neural information processing systems}, 2014, pp.
  2204--2212.

\bibitem{gregor2015draw}
Karol Gregor, Ivo Danihelka, Alex Graves, Danilo~Jimenez Rezende, and Daan
  Wierstra,
\newblock ``Draw: A recurrent neural network for image generation,''
\newblock {\em arXiv preprint arXiv:1502.04623}, 2015.

\bibitem{sonderby2015recurrent}
S{\o}ren~Kaae S{\o}nderby, Casper~Kaae S{\o}nderby, Lars Maal{\o}e, and Ole
  Winther,
\newblock ``Recurrent spatial transformer networks,''
\newblock {\em arXiv preprint arXiv:1509.05329}, 2015.

\bibitem{almahairi2016dynamic}
Amjad Almahairi, Nicolas Ballas, Tim Cooijmans, Yin Zheng, Hugo Larochelle, and
  Aaron Courville,
\newblock ``Dynamic capacity networks,''
\newblock in {\em International Conference on Machine Learning}, 2016, pp.
  2549--2558.

\bibitem{seer2014kinects}
S.~Seer, N.~Br{\"a}ndle, and C.~Ratti,
\newblock ``Kinects and human kinetics: A new approach for studying pedestrian
  behavior,''
\newblock {\em Transportation Research part C: Emerging technologies}, vol. 48,
  pp. 212--228, 2014.

\bibitem{corbetta2018large}
Alessandro Corbetta, Werner Kroneman, Maurice Donners, Antal Haans, Philip
  Ross, Marius Trouwborst, Sander Van~de Wijdeven, Martijn Hultermans, Dragan
  Sekulovski, Fedosja van~der Heijden, et~al.,
\newblock ``A large-scale real-life crowd steering experiment via arrow-like
  stimuli,''
\newblock {\em arXiv preprint arXiv:1806.09801}, 2018.

\end{thebibliography}

\end{document}